%% file: emnlp2020.tex
\documentclass[11pt,a4paper]{article}
\usepackage[hyperref]{emnlp2020}
\usepackage{times}
\usepackage{booktabs}
\usepackage{stfloats}
\usepackage{graphicx}
\usepackage{algorithm}
\usepackage[noend]{algpseudocode}
\usepackage{latexsym}
\usepackage{multirow}
\usepackage{wrapfig}
\usepackage{placeins}
\usepackage{scrextend}

\usepackage{microtype}

\aclfinalcopy 
\def\aclpaperid{2767} 

\newcommand{\greedy}[0]{\textsc{Greedy}}
\newcommand{\fixed}[0]{\textsc{Fixed}}
\newcommand{\varbeam}[0]{\textsc{Var-Batch}}
\newcommand{\varstream}[0]{\textsc{Var-Stream}}
\newcommand{\varfifo}[0]{\textsc{Var-Str-FIFO}}
\newcommand{\fixedours}[0]{\textsc{Fixed-Ours}}
\newcommand{\fixedstream}[0]{\textsc{Fixed-Stream}}
\newcommand{\fixedfairseq}[0]{\textsc{Fixed-Fairseq}}

\title{A Streaming Approach For Efficient Batched Beam Search} 

\author{Kevin Yang, Violet Yao, John DeNero, Dan Klein \\
\\
UC Berkeley\\
\\\{yangk,violetyao,denero,klein\}@berkeley.edu}

\date{}

\begin{document}
\maketitle
\begin{abstract}
We propose an efficient batching strategy for variable-length decoding on GPU architectures. During decoding, when candidates terminate or are pruned according to heuristics, our streaming approach periodically ``refills" the batch before proceeding with a selected subset of candidates. We apply our method to variable-width beam search on a state-of-the-art machine translation model. Our method decreases runtime by up to 71\% compared to a fixed-width beam search baseline and 17\% compared to a variable-width baseline, while matching baselines' BLEU.
Finally, experiments show that our method can speed up decoding in other domains, such as semantic and syntactic parsing. 
\end{abstract}

\input{intro}
\input{method}
\input{experiments}

\input{discussion}
\input{acknowledgments}
\bibliographystyle{acl_natbib}
\bibliography{emnlp2020}
\newpage
\appendix
\input{appendix}

\end{document}

%% file: intro.tex
\section{Introduction}

While inference is often cheap compared to training in modern neural models, one may need to run inference frequently or continually. Such is the case for online machine translation (MT) services: as far back as 2016, Google Translate already translated 100 billion words daily~\cite{turovsky_2016}. Large-scale inference is also required for methods such as iterative backtranslation and knowledge distillation to generate training data~\cite{hoang2018iterative,kim2016sequence}. For such high-throughput applications, it is useful to decrease inference cost. 

Meanwhile, we must preserve accuracy: beam search is slower than greedy decoding, but is nevertheless often preferred in MT. Not only is beam search usually more accurate than greedy search, but it also outputs a \textit{diverse set} of decodings, enabling reranking approaches to further improve accuracy~\cite{yee2019simple,ng2019facebook,charniak2005coarse,raymond2006discriminative}.  

However, it is challenging to optimize the performance of beam search for modern neural architectures. Unlike classical methods in sparse computation settings, modern neural methods typically operate in dense (batched) settings to leverage specialized hardware such as GPUs. 

In this work, we propose a streaming method to optimize GPU-batched variable-output-length decoding. Our method does not use a fixed batch during inference; instead, it continually ``refills" the batch after it finishes translating some fraction of the current batch. Our method then continues decoding on the remaining candidates in the batch, prioritizing those least expanded. 



We apply our method to variable-width beam search. For variable-output-length decoding even in batched settings, variable-width beam search often modestly decreases accuracy in exchange for substantial speedups over fixed-width beam search~\cite{freitag2017beam,wu2016google}. When decoding with Fairseq's state-of-the-art WMT'19 model~\cite{ng2019facebook}, our method further improves over the speed of baseline variable-width beam search: up to 16.5\% on a 32GB V100 GPU, without changing BLEU~\cite{papineni2002bleu}. Our approach also improves decoding efficiency in lightweight models for semantic and syntactic parsing.\footnote{Code available at \url{https://github.com/yangkevin2/emnlp2020-stream-beam-mt}.} In principle, our method can be applied to any task which sequentially processes variable-length data.

%% file: method.tex
\section{Background: Beam Search}

\begin{figure*}[t!]
    \centering
    \includegraphics[width=0.9\textwidth]{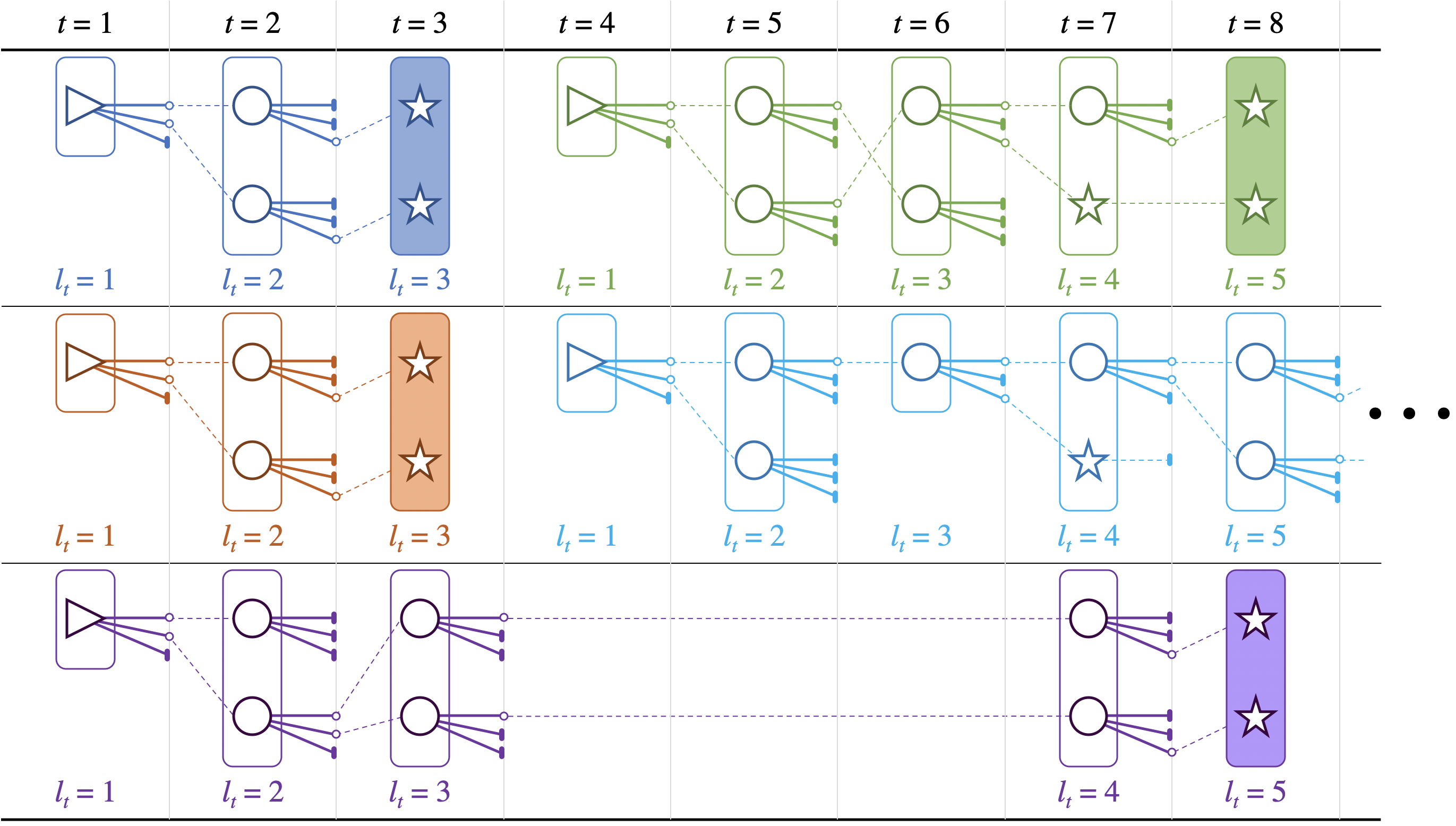}
    \caption{
    Illustration of our method \varstream{} for variable-width beam search with vocabulary size $|\mathcal{V}|=3$, beam width $k=2$, batch size $n=3$, refill threshold $\epsilon = \frac{1}{3}$. Each color corresponds to the beam for a single input. The rounded rectangles at each timestep are beams $\mathcal{H}(B_i^{l_t})$, while the shapes inside are individual candidates. Shaded beams represent the end of the search. The right-facing triangles indicate the initial candidate containing just the start token $w_{sos}$, circles denote an active (non-terminated) candidate, and stars denote a finalized candidate. Candidates become finalized after following the third (bottom-most) branch in an expansion, corresponding to the end token $w_{eos}$; they then undergo only no-op expansions thereafter. The first two rows of beams depict normal operation of variable-width beam search, including heuristic pruning in the light blue beam at $t=6$. The third row shows an important detail of our method: \varstream{} refills the batch after $t = 3$, when only $\epsilon n$ beams remain, and the remaining purple beam halts computation until the two newly added beams reach the same $l_t$. (This detail matters in transformer architectures; see Appendix \ref{sec:appendix_alt}.)
    }
    \label{fig:beam}
\end{figure*}

Given encoder $\mathcal{E}$ and decoder $\mathcal{D}$, our task is to convert inputs $\{x_1 \dots x_N\}$ into corresponding outputs $\{\bar{y}_1 \dots \bar{y}_N\}$, for data size $N$. For example, in machine translation, each $x_i$ is a source sentence consisting of a sequence of tokens and each $\bar{y}_i$ is a translation. We assume $\mathcal{D}(e_i, y_i)$ receives $e_i = \mathcal{E}(x_i)$ and a partial $y_i$ as input, constructing $\bar{y}_i$ one token at a time. 

One method of constructing $\bar{y}_i$ for a given $x_i$ is greedy search. Let $y_i^{l_t}$ be the in-construction \textit{candidate} with length $l_t = t$ at timestep $t$. We initialize $y_i^1$ as the start token $w_{sos}$, and at each timestep $t$ obtain $y_i^{l_t+1}$ by concatenating the maximum-probability token. We \textit{finalize} $y_i^{l_t}$ as $\bar{y}_i$ once we append the end token, or at some maximum length.


Previous work has found that greedy search often underperforms beam search in accuracy, and gains from non-greedy decoding have also been observed in many classical models~\cite{sutskever2014sequence,freitag2017beam,wilt2010comparison}. See the dark blue, green, and brown beams in Figure \ref{fig:beam} for normal operation of fixed-width beam search using beam width $k=2$. For each input $x_i$, fixed-width beam search tracks a length-$l_t$, width-$k$ \textit{beam} $B_i^{l_t}$ for each time $t$. $B_i^{l_t}$ contains $k$ length-$l_t$ candidates $y_{i1}^{l_t} \dots y_{ik}^{l_t}$ with maximum log-likelihood in order, denoted by the shapes inside the rounded rectangles (beams) in the figure. At each step, beam search considers all $k|\mathcal{V}|$ possible \textit{candidate expansions} (one-token extensions of existing candidates), where $\mathcal{V}$ is the vocabulary. The top $k$ expansions become the expanded beam $B_i^{l_t+1}$. Figure \ref{fig:beam} shows these expansions at each timestep for $|\mathcal{V}| = 3$, with active non-terminated candidates (circles) becoming finalized (stars) after following the bottom-most branch, corresponding to the end token $w_{eos}$. In the end, beam search yields $k$ finalized candidates $\bar{y}_{i1} \dots \bar{y}_{ik}$ compared to a single $\bar{y}_i$ in greedy search.

\textit{Variable-width} beam search reduces the computational cost of the above \textit{fixed-width} beam search by pruning the full beam $B_i^{l_t}$ using heuristics $\mathcal{H}$, for example at $t=6$ for the light blue beam in the figure. The width of the resulting pruned beam $\mathcal{H}(B_{i}^{l_t})$ is no longer always exactly equal to $k$, and may vary over time. 




\section{Streaming Variable-Length Decoding}


As an example, consider translating a batch of $n$ German sentences into English via a traditionally-batched variable-width beam search (e.g., \citet{freitag2017beam}) on a GPU. Henceforth we refer to this baseline as \varbeam{}.


An inefficiency results from decoding being inherently variable in length: After $t$ steps, we may have completed $m < n$ translations, but the last $n-m$ beams may take several more timesteps. For example, in Figure \ref{fig:beam}, our initial batch consists of the dark blue, brown, and purple beams. After the dark blue and brown beams terminate, we would still be stuck decoding the purple beam by itself.


The resulting GPU underutilization motivates our streaming approach \varstream{} applied to variable-width beam search (Figure \ref{fig:beam}). For batch size $n$, \varstream{} initially proceeds identically to \varbeam{}. But when the number of remaining beams drops below $\epsilon n$ for some constant $\epsilon \in (0, 1)$, \varstream{} encodes a new batch of inputs $x$ to ``refill" its batch to size $n$.\footnote{Our method is relatively insensitive to $\epsilon$ (Appendix \ref{sec:appendix_apmceps}).} This occurs at $t=4$ in Figure \ref{fig:beam}, where we refill the batch using the green and light blue beams. 

Note the active beams are no longer of equal length $l_t = t$ for every beam after refilling. At each subsequent $t$, \varstream{} only expands beams $\mathcal{H}(B_{i}^{l_t})$ with minimal $l_t$; in particular, the purple beam in Figure \ref{fig:beam} pauses computation at $t=4$.\footnote{As very long translations could get ``stuck" in the batch, one can periodically finish computation on all remaining beams in the batch if latency is a concern in addition to throughput.} When decoding with state-of-the-art transformer architectures for MT, it is advantageous to expand only beams with minimal $l_t$ at each step, because self-attention causes steps at higher $l_t$ to be more expensive; see Appendix \ref{sec:appendix_alt}. (For RNN-based architectures, it may be faster to expand all active beams at each step.) 

We emphasize that \varstream{} is an implementation optimization, exactly matching the output of \varbeam{}. Full details in Algorithm \ref{alg:algorithm}.

When the memory bottleneck is partially the decoding process itself rather than caching the input encodings $\mathcal{E}(x_i)$ or beams $\mathcal{H}(B_{i}^{l_t})$, \varstream{} can cache additional encodings and beams on GPU. At each $t$, \varstream{} then selects beams up to some limit on total beam width, filling GPU capacity even in the case of variable-width beams. This batching constraint addresses a second inefficiency in GPU utilization: the widths of the pruned beams $\mathcal{H}(B_{i}^{l_t})$ may vary over time. We exploit this in semantic (Sec. \ref{semantic}) and syntactic parsing (Sec. \ref{syntactic}).

\begin{algorithm}[t] 
    \caption{\varstream{}}
    \label{alg:algorithm}
    
    \textbf{Input:} inputs $X = \{x_1, \dots x_N\}$, model $(\mathcal{E}, \mathcal{D})$, batch size $n$, refill threshold $\epsilon$, beam width $k$, pruning heuristics $\mathcal{H}$
    
    \begin{algorithmic}[1]
        \Procedure{Decode}{$X, \mathcal{E}, \mathcal{D}, n, \epsilon, k, \mathcal{H}$}
            \State \textit{\# initialize encodings, beams, final outputs}
            \State $E, \mathcal{B}, Y = [\;], [\;], [\;]$ 
            \State \textit{\# initialize index counter}
            \State $c=1$
            \While {$c \leq N$ or $|\mathcal{B}| > 0$}
                \If{$c \leq N$ and $|\mathcal{B}| \leq \epsilon n$}
                    \State \textit{\# refill batch by $m = n(1-\epsilon)$}
                    \State $E = E + [\mathcal{E}(x_c) \dots \mathcal{E}(x_{c+m-1})]$
                    \State $\mathcal{B} = \mathcal{B} + [w_{sos}] \times m$
                    \State $Y = Y + [\;] \times m$
                    \State $c = c + m$
                \EndIf
                \State Select $E_s \in E, \mathcal{B}_s \in \mathcal{B}$ with min $l_t$
                \For {$(e_i, \mathcal{H}(B_{i}^{l_t})) \in (E_s, \mathcal{B}_s)$}
                \For {$y_{ij}^{l^t} \in \mathcal{H}(B_{i}^{l_t})$}
                \State Compute expansion $\mathcal{D}(e_i, y_{ij}^{l^t})$
                \EndFor
                \State Update $\mathcal{H}(B_{i}^{l_t})$ to $\mathcal{H}(B_{i}^{l_t+1})$
                \State Add finalized candidates to $Y[i]$
                \EndFor
                \State Remove terminated beams from $E, \mathcal{B}$
            \EndWhile
            \State\Return{$Y$} 
        \EndProcedure
    \end{algorithmic}
\end{algorithm}

%% file: experiments.tex
\section{Experiments}

\begin{table}[!t]
\centering
\addtolength{\tabcolsep}{-3pt}
\begin{tabular}{@{\hspace{-1ex}}l@{}lrr@{}}
\addlinespace[-\aboverulesep] 
 \cmidrule[\heavyrulewidth]{2-4}
& \textbf{Method}     & \multicolumn{1}{l}{\textbf{BLEU}} & \multicolumn{1}{l}{\textbf{Wall Clock (s)}} \\ \cmidrule{2-4}
\multicolumn{1}{c}{\textbf{\textit{De-En}}} & \greedy{}              & 48.18\phantom{*}                              & 39.01 $\pm$ \phantom{2}0.14                               \\ 
\multicolumn{1}{c}{\textbf{\textit{k=50}}} & \fixed{}               & 49.57\phantom{*}                              & 891.53 $\pm$ \phantom{2}0.77                              \\ 
\multicolumn{1}{c}{\textbf{\textit{32GB}}}& \varbeam{}           & 49.59*                             & 308.15 $\pm$ \phantom{2}3.48                              \\ 
& \textit{\varstream{}} &    49.59*                  & 257.20 $\pm$ \phantom{2}0.54                     \\ \cmidrule[\heavyrulewidth]{2-4}
& \textbf{Method}     & \multicolumn{1}{l}{\textbf{BLEU}} & \multicolumn{1}{l}{\textbf{Wall Clock (s)}} \\ \cmidrule{2-4}
\multicolumn{1}{c}{\textbf{\textit{Ru-En}}}& \greedy{}              & 37.84\phantom{*}                              & 42.09 $\pm$ \phantom{2}1.33                               \\ 
\multicolumn{1}{c}{\textbf{\textit{k=50}}} & \fixed{}               & 38.98\phantom{*}                              & 893.23 $\pm$ \phantom{2}1.23                              \\ 
\multicolumn{1}{c}{\textbf{\textit{32GB}}} & \varbeam{}           & 39.04*                              & 399.98 $\pm$ \phantom{2}1.25                              \\ 
& \textit{\varstream{}} &    39.04*                  & 342.18 $\pm$ \phantom{2}1.28\\   
\cmidrule[\heavyrulewidth]{2-4}
& \textbf{Method}     & \multicolumn{1}{l}{\textbf{BLEU}} & \multicolumn{1}{l}{\textbf{Wall Clock (s)}} \\ 
\cmidrule{2-4}
\multicolumn{1}{c}{\textbf{\textit{De-En}}} & \greedy{}              & 48.18\phantom{*}                              & 46.41 $\pm$ \phantom{2}0.23                               \\ 
\multicolumn{1}{c}{\textbf{\textit{k=5}}} & \fixed{}               & 49.42\phantom{*}                              & 102.25 $\pm$ \phantom{2}0.58                            \\ 
 \multicolumn{1}{c}{\textbf{\textit{32GB}}}& \varbeam{}           & 49.46*                              & 114.18 $\pm$ \phantom{2}0.17                             \\ 
& \textit{\varstream{}} &   49.46*                   & 92.39 $\pm$ \phantom{2}1.05                     \\ \cmidrule[\heavyrulewidth]{2-4}
& \textbf{Method}     & \multicolumn{1}{l}{\textbf{BLEU}} & \multicolumn{1}{l}{\textbf{Wall Clock (s)}} \\ \cmidrule{2-4}
\multicolumn{1}{c}{\textbf{\textit{Ru-En}}}& \greedy{}              & 37.84\phantom{*}                              & 42.09 $\pm$ \phantom{2}1.33                               \\ 
\multicolumn{1}{c}{\textbf{\textit{k=5}}} & \fixed{}               & 38.83\phantom{*}                              & 103.59 $\pm$ \phantom{2}0.34                              \\ 
\multicolumn{1}{c}{\textbf{\textit{32GB}}} & \varbeam{}           & 39.03*                              & 130.01 $\pm$ \phantom{2}2.33                              \\ 
& \textit{\varstream{}} &    39.03*                  & 95.45 $\pm$ \phantom{2}0.21\\   
\cmidrule[\heavyrulewidth]{2-4}
& \textbf{Method}     & \multicolumn{1}{l}{\textbf{BLEU}} & \multicolumn{1}{l}{\textbf{Wall Clock (s)}} \\ 
\cmidrule{2-4}
\multicolumn{1}{c}{\textbf{\textit{De-En}}} & \greedy{}              & 48.18\phantom{*}                              & 46.41 $\pm$ \phantom{2}0.23                               \\ 
\multicolumn{1}{c}{\textbf{\textit{k=50}}} & \fixed{}               & 49.57\phantom{*}                              & 2072.86 $\pm$ 23.18                            \\ 
 \multicolumn{1}{c}{\textbf{\textit{16GB}}}& \varbeam{}           & 49.59*                              & 645.70 $\pm$ 17.49                             \\ 
& \textit{\varstream{}} &   49.59*                   & 606.17 $\pm$ \phantom{2}4.96                     \\ \cmidrule[\heavyrulewidth]{2-4}
& \textbf{Method}     & \multicolumn{1}{l}{\textbf{BLEU}} & \multicolumn{1}{l}{\textbf{Wall Clock (s)}} \\ \cmidrule{2-4}
\multicolumn{1}{c}{\textbf{\textit{Ru-En}}}& \greedy{}              & 37.84\phantom{*}                              & 52.26 $\pm$ \phantom{2}0.50                               \\ 
\multicolumn{1}{c}{\textbf{\textit{k=50}}} & \fixed{}               & 38.98\phantom{*}                              & 2155.95 $\pm$ 58.47                            \\ 
\multicolumn{1}{c}{\textbf{\textit{16GB}}} & \varbeam{}           & 39.04*                              & 852.93 $\pm$ \phantom{2}9.11                              \\ 
& \textit{\varstream{}} &   39.04*                   & 803.72 $\pm$ 15.94                    \\ \cmidrule[\heavyrulewidth]{2-4}
\end{tabular}
\addtolength{\tabcolsep}{3pt}
\caption{Top-1 BLEU and wall clock times for machine translation. Our method \varstream{} is substantially faster than \varbeam{} ($14$-$17\%$ for $k=50$ on 32GB, $19$-$27\%$ for $k=5$ on 32GB, $6\%$ for $k=50$ on 16GB) and \fixed{} ($62$-$71\%$ for $k=50$, $8$-$10\%$ for $k=5$), while preserving high BLEU. *Our rules for finalizing candidates during decoding differ slightly from Fairseq, resulting in equal or higher BLEU for \varbeam{} and \varstream{} compared to \fixed{}. Adapting our implementation to fixed-width beam search is slower but yields higher BLEU (Appendix \ref{sec:appendix_impl_diff}).\footref{note1}}
\label{tab:mt}
\end{table}

We apply \varstream{} to variable-width beam search in machine translation, semantic parsing, and syntactic parsing. We use the absolute threshold and max candidates heuristics of \citet{freitag2017beam} as $\mathcal{H}$, modifying only the heuristic hyperparameters for each domain based on a development set. The absolute threshold heuristic prunes candidates $y_{ij}^{l_t}$ whose log-probabilities fall short of the best candidate $y_{i1}^{l_t}$'s by some threshold $\delta$, i.e. $\log P(y_{ij}^{l_t}) < \log P(y_{i1}^{l_t}) - \delta$. The max candidates heuristic prevents the search from selecting more than $M < k$ length-$l_t+1$ candidates originating from the same length-$l_t$ candidate at each step $t$.

In each domain we compare four methods:
\begin{enumerate}
\itemsep0em
    \item \textit{\greedy{}}, a greedy search, 
    \item \textit{\fixed{}}, a fixed-width beam search,
    \item \textit{\varbeam{}}, a batched variable-width beam search, and
    \item \textit{\varstream{}}, our streaming method.
\end{enumerate}

We sort and bucket inputs by length for batching.

\subsection{Machine Translation}

We evaluate on the transformer architecture implemented in Fairseq~\cite{ott2019fairseq}, which scored highest on several tracks of WMT'19 \cite{barrault2019findings}. For our main experiments, we run German-English and Russian-English translation on newstest2018 using an ensemble of $5$ models with $k = 50$, matching the setup of \citet{ng2019facebook} but without reranking. As smaller beam sizes are also common in practice, we evaluate with $k=5$ as well. Our \greedy{} and \fixed{} baselines are Fairseq's implementation, while \varbeam{} and \varstream{} are our own. For all methods, we evaluate 5 runs on a 32GB V100 GPU. For $k=50$, we also run on a 16GB V100 GPU, noting that 32GB is likely more realistic in a production setting. We choose batch size to saturate the GPU, using $\epsilon = \frac{1}{6}$ for \varstream{}, with pruning heuristics $\delta = 1.5, M=5$. Appendix \ref{sec:appendix_apmceps} details hyperparameter choices. 

\begin{table*}[t]
\centering
\addtolength{\tabcolsep}{-1pt}

\begin{tabular}{@{}lcccccccc@{}}
\toprule
                    & \multicolumn{4}{c}{\textit{\textbf{Semantic Parsing (ATIS)}}}                                                                                                     & \multicolumn{4}{c}{\textit{\textbf{Syntactic Parsing (Penn Treebank)}}}                                                                                           \\ \cmidrule(lr){2-5} \cmidrule(lr){6-9}
\textbf{Method}     & \multicolumn{1}{l}{\textbf{F1}} & \multicolumn{1}{l}{\textbf{Oracle}} & \multicolumn{1}{l}{\textbf{Time (s)}} & \multicolumn{1}{l}{\textbf{Exp. / Step}} & \multicolumn{1}{l}{\textbf{F1}} & \multicolumn{1}{l}{\textbf{Oracle}} & \multicolumn{1}{l}{\textbf{Time (s)}} & \multicolumn{1}{l}{\textbf{Exp. / Step}} \\ \hline
\greedy{}              & 86.4                            & 86.4                                & 1.4 $\pm$ 0.0                           & 46.5                                     & 91.3                            & 91.3                                & \phantom{2}22.3 $\pm$ 0.2                          & 86.7                                     \\
\fixed{}               & 86.6                            & 91.2                                & 7.7 $\pm$ 0.1                           & 48.0                                       & 91.2                            & 94.0                                  & 224.2 $\pm$ 2.1                         & 97.1                                     \\
\varbeam{}           & 86.6                            & 90.2                                & 8.2 $\pm$ 0.2                           & 16.9                                     & 91.2                            & 93.8                                & 235.6 $\pm$ 1.8                         & 48.2                                     \\
\textit{\varstream{}} & 86.6                            & 90.2                                & 6.3 $\pm$ 0.1                           & 72.1                                     & 91.2                            & 93.8                                & 220.5 $\pm$ 2.5                         & 95.7      \\ \bottomrule
\end{tabular}
\addtolength{\tabcolsep}{1pt}

\caption{Top-1 and oracle reranking F1, wall clock (avg. 5 runs), and average candidate expansions per timestep (i.e., total candidate expansions divided by total decoding timesteps) for semantic parsing on ATIS and syntactic parsing on the Penn Treebank (PTB). Theoretical maximum efficiency under our batching constraint is 100 expansions per step for both tasks. \varstream{} achieves substantially higher expansions per step than other methods on ATIS. On PTB, \fixed{} achieves near-perfect efficiency because all $\bar{y}_{ij}$ for a given $x_i$ have the same length. But comparing variable-width beam searches, \varstream{} is much more efficient with batch capacity than \varbeam{}.}
\label{tab:parsing}
\end{table*}

As shown in Table \ref{tab:mt}, on both GPU settings and on both languages, \greedy{} is fastest, but suffers heavily in BLEU. Our \varstream{} is the fastest beam-based search, and matches the BLEU of the beam search baselines. Compared to \varbeam{}, \varstream{} is faster by $14$-$17\%$ when using $k=50$ on the 32GB GPU, and by $19$-$27\%$ when $k=5$. \varstream{} also remains $6\%$ faster when using $k=50$ on the 16GB GPU where overhead is higher. \varbeam{} and \varstream{} match the BLEU of \fixed{} while being 2-3 times faster when using beam size 50, confirming the speedups from \varbeam{} over \fixed{} in e.g., \citet{freitag2017beam}. \fixed{} is more competitive when $k=5$ because the potential for heuristic beam pruning is much more limited; moreover, our implementations of \varstream{} and \varbeam{} somewhat understate both speedups and BLEU cost compared to \fixed{} due to an implementation difference with Fairseq (Appendix \ref{sec:appendix_impl_diff}).\footnote{\label{note1}Essentially, we allow finalized candidates to fall off the beam if we find enough other higher-likelihood candidates. See e.g., the star at $t=7$ in the light blue beam in Figure \ref{fig:beam}. Fairseq does not allow this.} Thus \varbeam{} becomes slower than \fixed{} when $k=5$. Nevertheless, \varstream{} remains the fastest in this scenario by $8$-$10\%$.

\subsection{Semantic Parsing}\label{semantic}

To explore our method's domain applicability, we experiment with semantic parsing using the seq2seq model of \citet{dong2016language}. This lightweight model is no longer state of the art, but its decoding is representative of more recent architectures~\cite{suhr2018learning,yin2018tranx,lin2019grammar}. We use the ATIS flight-booking dataset~\cite{dahl1994expanding}, setting $n=k=\delta = 10, M = 3$. Due to the small dataset and model, our batching constraint is more theoretical: we constrain each method to expand at most $nk = 100$ candidates per timestep (i.e., total beam width), instead of simply saturating the GPU.\footnote{Due to caching additional encodings and beams, \varstream{} uses more GPU memory in this idealized setting.}

As shown by the expansions per step in Table \ref{tab:parsing}, \varstream{} uses the batch capacity of $100$ most efficiently.
Thus \varstream{} is faster than both \varbeam{} and \fixed{}, despite overhead which is exacerbated in a small model. The speedup is larger on the JOBS and GEO datasets~\cite{zettlemoyer2012learning} (Appendix \ref{sec:appendix_semparseexp}). While all methods achieve similar top-1 F1, oracle F1 (using an oracle to ``rerank" all outputs $\bar{y}_{ij}$) highlights the benefit of producing a diverse set of translations.

\subsection{Syntactic Parsing}\label{syntactic}

We also experiment with the lightweight shift-reduce constituency parser of \citet{cross2016span} on the Penn Treebank \cite{marcus1993building}. 
This task and model differ from our previous setups in that for a given input $x_i$, all valid parses $\bar{y}_{ij}$ have exactly the same length. When inputs are bucketed by length, this removes the variable-output-length inefficiency for traditional batching: we cannot get stuck finishing a small fraction of beams when the rest of the batch is done. Thus, this task isolates the effect of \varstream{} using batch capacity more efficiently in the case of variable-width beams. We use the same computational constraint as in semantic parsing, with $n = k = 10,\delta=2.5, M=3$. 


As all $\bar{y}_{ij}$ have equal length for a given $x_i$, \fixed{} already achieves near-perfect efficiency in expansions per step (Table \ref{tab:parsing}). 
Combined with the impact of overhead in this older (smaller) model, \varstream{} is not substantially faster than \fixed{} in this setting. However, when comparing variable-width beam searches where efficient batching is more difficult, we observe that \varstream{} doubles \varbeam{} in expansions per step.






%% file: discussion.tex
\section{Discussion}

In this work, we have proposed a streaming method for variable-length decoding to improve GPU utilization, resulting in cheaper inference. Applied to a state-of-the-art machine translation model, our method yields substantial speed improvements compared to traditionally-batched variable-width beam search. We also apply our method to both semantic and syntactic parsing, demonstrating our method's broader applicability to tasks that process variable-output-length data in a sequential manner. 


%% file: acknowledgments.tex
\section*{Acknowledgments}

We thank Steven Cao, Daniel Fried, Nikita Kitaev, Kevin Lin, Mitchell Stern, Kyle Swanson, Ruiqi Zhong, and the three anonymous reviewers for their helpful comments and feedback, which helped us to greatly improve the paper. This work was supported by Berkeley AI Research, DARPA through the Learning with Less Labeling (LwLL) grant, and the NSF through a fellowship to the first author.

%% file: appendix.tex
\section{Appendices}
\label{sec:appendix}

\FloatBarrier

\subsection{Implementation Differences Compared to Fairseq}
\label{sec:appendix_impl_diff}

In our main machine translation experiments, the \fixed{} baseline is Fairseq's implementation. Running our own beam search implementation---the basis of \varbeam{} and \varstream{}---with a fixed beam width differs from Fairseq's implementation as follows. In our implementation, henceforth \fixedours{}, terminated candidates $y_{ij}^{l_{t_0}}$ with $i > 1$ are kept on the beam, added to our list of final outputs only if they become the top candidate $y_{i1}^{l_{t_1}}$ in the beam at a subsequent step $t_1$. Fairseq instead immediately adds $y_{ij}^{l_{t_0}}$ to the list of final outputs at time $t_0$. The difference is that $y_{ij}^{l_{t_0}}$ may be removed from the beam at time $t > t_0$ if we later find multiple terminated candidates originating from a higher-probability beam $y_{ij'}^{l_{t_0}}$ for $j' < j$, e.g. between $t=7$ and $t=8$ in the light blue beam in Figure \ref{fig:beam}.

\fixedours{} is slower than Fairseq's implementation. However, while the two implementations achieve more similar BLEU on the development set, \fixedours{} achieves higher BLEU on the test set (49.75 vs 49.57 on De-En and 39.19 vs 38.98 on Ru-En). See Table \ref{tab:our_beam} for De-En experiment details.

\begin{table}[]
\begin{tabular}{@{}lrr@{}}
\toprule
\textbf{Method}     & \multicolumn{1}{l}{\textbf{BLEU}} & \multicolumn{1}{l}{\textbf{Wall Clock (s)}} \\\hline
\fixedfairseq{}     & 49.57                                   & 891.53 $\pm$ 0.77                             \\
\textit{\fixedours{}}        & 49.75                                   &  1280.59 $\pm$ 5.34                                           \\
\textit{\fixedstream{}} & 49.75                                   & 1004.18 $\pm$ 6.82     \\    
\bottomrule
\end{tabular}
\caption{De-En translation experiments test set (newstest2018) on 32GB Nvidia V100 using different implementations of fixed-size beam search. \fixedfairseq{} is the \fixed{} baseline in the main paper, while \fixedours{} is our implementation of fixed-size beam search. \fixedstream{} is a streaming implementation with $\epsilon=\frac{1}{6}$; \fixedours{} corresponds to $\epsilon = 0$. \fixedstream{} improves over \fixedours{} in wall clock, but is still slower than \fixedfairseq{}, although it achieves higher BLEU. }
\label{tab:our_beam}
\end{table}

For completeness, we also present results in Table \ref{tab:our_beam} for \fixedstream{}, our streaming implementation adapted to fixed-size beam search on newstest2018 on the 32GB Nvidia V100, with $k=50$ as in the \fixed{} baseline. We keep the $\epsilon=\frac{1}{6}$ hyperparameter. \fixedstream{} is significantly faster than \fixedours{}, demonstrating that our streaming method can also speed up fixed-size beam search. However, \fixedstream{} is slower than Fairseq's implementation, although it outperforms Fairseq. It is possible that our implementation is less optimized, but we do not formally claim this.

\FloatBarrier

\subsection{Alternative Method Analysis}\label{sec:appendix_alt}

\begin{table}[]
\centering
\begin{tabular}{@{}lrr@{}}
\toprule
\textbf{Method} & \multicolumn{1}{l}{\textbf{Wall Clock (s)}} & \multicolumn{1}{l}{\textbf{Timesteps}} \\ \hline
\varbeam{}       & 308.15 $\pm$ 3.48                              & 2007                                    \\ 
\textit{\varstream{}}      & 257.20 $\pm$ 0.54                              & 2180                                    \\ 
\textit{\varfifo{}} & 343.10 $\pm$ 0.41                              & 1538                                    \\ \bottomrule
\end{tabular}
\caption{Comparison of variable-size beam search methods on De-En translation of newstest2018 on 32GB Nvidia V100 GPU. All methods achieve equal BLEU and expand the same number of candidates. \varfifo{} uses the fewest timesteps but takes the most time due to the transformer architecture requiring more time per step at high values of $l_t$.}
\label{tab:ablation}
\end{table}

\begin{figure}
    \centering
    \includegraphics[width=0.5\textwidth]{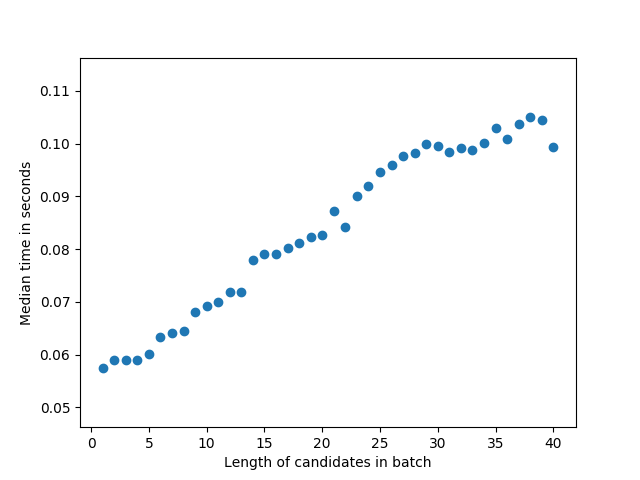}
    \caption{Median time for decoding steps at values of $l_t$ up to 40, for a single run of De-En translation on the 32GB GPU with $k=50$. The time taken increases roughly linearly with $l_t$ due to self-attention in the transformer.}
    \label{fig:lt_time}
\end{figure}

We briefly analyze an alternative streaming method to our proposed \varstream{}, which we label  \varfifo{}. At each decoding timestep $t$, instead of selecting only the beams with minimal $l_t$ as in \varstream{}, \varfifo{} selects beams up to its batch capacity starting with the beam of maximal $l_t$. In Figure \ref{fig:beam}, this corresponds to not pausing computation for the purple beam. This is intuitively appealing and has potential advantages: as shown in Table \ref{tab:ablation}, unlike \varstream{} which uses slightly more timesteps than \varbeam{} due to using a slightly smaller effective batch size, \varfifo{} significantly reduces the number of timesteps required for decoding in the De-En translation task. Yet \varfifo{} is significantly slower than both \varstream{} and \varbeam{}. This is due to Fairseq's architecture, a transformer reliant on decoder self-attention, causing decoding timesteps with longer $l_t$ to be more expensive (Figure \ref{fig:lt_time}). \varfifo{} suffers because it must pad all selected beams' lengths up to the maximum $l_t$ among those selected.

The difference between \varstream{} and \varfifo{} demonstrates that selecting the correct beams to expand during decoding timesteps can be highly impactful on speed, illustrating a new axis of optimization made possible by streaming. While \varstream{} is superior for the transformers used in state-of-the-art machine translation, we hypothesize that \varfifo{} may be preferred in other applications, especially in RNN architectures which do not use self-attention.

\FloatBarrier

\subsection{Experiment Details and Hyperparameters}\label{sec:appendix_apmceps}

We provide details on our setup. All code is written in Pytorch~\cite{paszke2017automatic}. For hardware, for our 32GB and 16GB Nvidia V100 experiments, we use p3dn.24xlarge and p3.2xlarge instances respectively on AWS. Experiments are conducted serially with no other computation on the instance. Due to some variance between instances, all experiments within a single comparable group (e.g., all methods' runs for beam size 50 on a 32GB GPU) are conducted on the same instance. 

Additionally, we specify an implementation detail: for the absolute threshold heuristic $\delta$, we note that $y_{i1}^{l_t}$ may be a terminated candidate from a previous timestep.

For heuristic hyperparameters, in all domains we choose pruning heuristics to approximately match the performance of \fixed{}, based on the development set (newstest2017 for machine translation, and the ATIS and Penn Treebank development sets in semantic and syntactic parsing respectively). As usual, in heuristic selection, there is a tradeoff between time and performance, which we explore here in the machine translation domain. 

We run \fixed{} and \varstream{} on the development set (newstest2017) for German-English translation using the 32GB Nvidia V100, using our main paper heuristics $\delta=1.5,M=5,\epsilon=\frac{1}{6}$. We additionally run versions where we individually tweak each heuristic, using $\delta=0.5,\delta=2.5,M=3,M=10,\epsilon=\frac{1}{12},\epsilon=\frac{1}{4}$.

\begin{table*}[t]
\centering
\begin{tabular}{@{}lrrrrr@{}}
\toprule
\textbf{Method} & \multicolumn{1}{l}{\textbf{Top 1 BLEU}} & \multicolumn{1}{l}{\textbf{Wall Clock (s)}} & \multicolumn{1}{l}{\textbf{Timesteps}} & \multicolumn{1}{l}{\textbf{Cand. Exp.}} & \multicolumn{1}{l}{\textbf{Exp. / Step}} \\ \midrule
\fixed{}           & 42.71                                   & 921.57 $\pm$ 2.95                             & 6620                                   & 3967200                                 & 599                                      \\
\varstream{}      & 42.71                                   & 259.78 $\pm$ 2.52                             & 2261                                   & 651786                                  & 288                                      \\
$\delta=0.5$          & 42.49                                   & 104.28 $\pm$ 1.35                             & 860                                    & 131108                                  & 152                                      \\
$\delta=2.5$          & 42.72                                   & 558.71 $\pm$ 5.49                             & 4325                                   & 1788047                                 & 413                                      \\
$M=3$            & 42.7                                    & 248.45 $\pm$ 2.66                             & 2070                                   & 625099                                  & 302                                      \\
$M=10$           & 42.71                                   & 268.62 $\pm$ 3.91                             & 2543                                   & 655300                                  & 258                                      \\
$\epsilon=\frac{1}{12}$        & 42.71                                   & 255.03 $\pm$ 0.49                             & 2479                                   & 651786                                  & 263                                      \\
$\epsilon=\frac{1}{4}$         & 42.71                                   & 264.73 $\pm$ 1.05                             & 2352                                   & 651786                                  & 277                                      \\ \bottomrule
\end{tabular}
\caption{Exploration of effect of different hyperparameter choices on \varstream{} performance on De-En development set (newstest2017). Top 1 BLEU, wall clock (average of 5 runs), total decoding timesteps, total candidate expansions, and average candidate expansions per step for several methods. All experiments on 32GB Nvidia V100. Larger values of $\delta$ and $M$ result in a method closer to fixed-width beam search, which tends to increase BLEU while taking more time due to needing more candidate expansions. Efficiency in expansions per step generally increases as the method approaches fixed-width beam search, as variable-width beams are more difficult to pack efficiently. Nevertheless, for the purpose of wall-clock time, this effect is outweighed by the vastly larger number of expansions required by fixed-width beam search.}
\label{tab:ablation}
\end{table*}

Both BLEU scores and computation time overall increase with $\delta$ and $M$ (Table \ref{tab:ablation}).

$\epsilon$ does not affect BLEU. Larger $\epsilon$ means we run fewer timesteps at high $l_t$, but our batch refills are smaller. At least in this machine translation setting, the effect of changing $\epsilon$ is typically a few seconds, indicating that our method is not overly sensitive to this hyperparameter choice as long as $\epsilon$ is small. (Note we re-adjusted batch sizes in multiples of 64 to saturate the GPU for each ablation.) 

During initial hyperparameter selection, we ran $\epsilon = \frac{1}{6}, \frac{1}{3}$ and $\frac{1}{2}$. For $M$ we tested $3, 5$, and $10$ and for $\delta$ we tested $1.5, 2.5, 5$ and $10$ based on manual tuning with single runs. Note that BLEU and F1 scores do not change with multiple trials, as we do not retrain models. Meanwhile, runtimes generally have fairly small standard deviation (see all tables), so we did not heavily optimize. Overall, the speedups enabled by \varstream{} over baselines are relatively insensitive to heuristic hyperparameters. 

\FloatBarrier

\subsection{Additional Semantic Parsing Experiments}\label{sec:appendix_semparseexp}

\begin{table*}[t]
\centering
\begin{tabular}{@{}lrrrrrr@{}}
\toprule
\textbf{Method} & \multicolumn{1}{l}{\textbf{F1}} & \multicolumn{1}{l}{\textbf{Oracle}} & \multicolumn{1}{l}{\textbf{Time (s)}} & \multicolumn{1}{l}{\textbf{Timesteps}} & \multicolumn{1}{l}{\textbf{Cand. Exp.}} & \multicolumn{1}{l}{\textbf{Exp. / Step}} \\ \midrule
\greedy{}          & 87.1                            & 87.1                                & 0.67 $\pm$ 0.02                         & 136                                    & 2715                                    & 20.0                                       \\
\fixed{}           & 87.1                            & 90.0                                  & 2.36 $\pm$ 0.04                         & 705                                    & 27781                                   & 39.4                                     \\
\varbeam{}       & 87.1                            & 89.3                                & 2.29 $\pm$ 0.02                         & 585                                    & 5071                                    & 8.7                                     \\
\varstream{}      & 87.1                            & 89.3                                & 1.47 $\pm$ 0.01                         & 126                                    & 5071                                    & 40.2           
\\ \bottomrule
\end{tabular}
\caption{Semantic parsing experiments on JOBS dataset of job listings. Top-1 F1, oracle reranking F1, wall clock average of 5 runs, total decoding timesteps, total candidate expansions, and average expansions per timestep. Although \varstream{} is not much more efficient than \fixed{} in expansions per timestep, it requires many fewer total expansions and is thus faster. Meanwhile, \varstream{} is several times more efficient than \varbeam{}. However, the variable beam searches suffer slightly in oracle F1 compared to \fixed{}, while still remaining above \greedy{}. }
\label{tab:jobs}
\end{table*}


\begin{table*}[]
\centering
\begin{tabular}{@{}lrrrrrr@{}}
\toprule
\textbf{Method} & \multicolumn{1}{l}{\textbf{F1}} & \multicolumn{1}{l}{\textbf{Oracle}} & \multicolumn{1}{l}{\textbf{Time (s)}} & \multicolumn{1}{l}{\textbf{Timesteps}} & \multicolumn{1}{l}{\textbf{Cand. Exp.}} & \multicolumn{1}{l}{\textbf{Exp. / Step}} \\ \midrule
\greedy{}          & 82.5                            & 82.5                                & 0.96 $\pm$ 0.02                         & 138                                    & 5492                                    & 39.8                                     \\
\fixed{}           & 82.5                            & 89.3                                & 4.54 $\pm$ 0.09                         & 1469                                   & 57550                                   & 39.2                                     \\
\varbeam{}       & 82.5                            & 89.6                                & 4.77 $\pm$ 0.05                         & 1344                                   & 14154                                   & 10.5                                     \\
\varstream{}      & 82.5                            & 89.6                                & 2.95 $\pm$ 0.04                         & 248                                    & 14154                                   & 57.1                              \\ \bottomrule      
\end{tabular}
\caption{Semantic parsing experiments on GEO dataset of geographical queries. \varstream{} is substantially more efficient than both \fixed{} and \varbeam{} in expansions per timestep, and this is reflected in the wall clock time.}
\label{tab:geo}
\end{table*}

In Tables \ref{tab:jobs} and \ref{tab:geo}, we present results from applying our method to the JOBS and GEO datasets. We use the same hyperparameters and heuristics as for ATIS, and operate under the same candidate-expansion constraint. \varstream{} is substantially faster than Fixed and \varbeam{} under this setting. 

\FloatBarrier

\subsection{Dataset Details}

\subsubsection{Machine Translation}

Evaluation datasets (newstest2018 and newstest2017) are available at \url{http://www.statmt.org/wmt19/translation-task.html}. newstest2018 contains 2998 and 3000 examples for De-En and Ru-En respectively, while newstest2017 contains 3004 and 3001. 

\subsubsection{Semantic Parsing}

Datasets can be obtained by running the data scripts at \url{https://github.com/Alex-Fabbri/lang2logic-PyTorch}, which re-implements \citet{dong2016language} in PyTorch. We use \citet{dong2016language}'s training, development (for ATIS), and test sets. ATIS, JOBS, and GEO contain 5410, 640, and 880 examples respectively. 

\subsubsection{Syntactic Parsing}

The Penn Treebank dataset and splits are available at \url{https://github.com/jhcross/span-parser}. The training, data, and test splits are the standard Penn Treebank splits (sections 2-21 for training, 22 for development, and 23 for test, containing 39832, 1700, and 2416 examples respectively). 